\documentclass[journal]{IEEEtran}
\usepackage{amsmath,amsfonts}
\usepackage{algorithmic}
\usepackage{algorithm}
\usepackage{array}
\usepackage[caption=false,font=normalsize,labelfont=sf,textfont=sf]{subfig}
\usepackage{textcomp}
\usepackage{stfloats}
\usepackage{url}
\usepackage{verbatim}
\usepackage{graphicx}
\usepackage{cite}
\usepackage[colorlinks,linkcolor=blue]{hyperref}
\usepackage{epstopdf}
\usepackage{multirow}
\usepackage{booktabs}
\usepackage{silence}
\begin{document}

\title{Deep Learning-based 3D Point Cloud Classification: A Systematic Survey and Outlook}

\author{Huang Zhang\dag, Changshuo Wang\dag, Shengwei Tian*, Baoli Lu, Liping Zhang, Xin Ning, Xiao Bai

\thanks{Huang Zhang and Shengwei Tian are with the School of Software, Xinjiang University, Xinjiang 830000, China (E-mail: zhhh1998@outlook.com; 357348035@qq.com).}

\thanks{Changshuo Wang is with the Institute of Semiconductors, Chinese Academy of Sciences, Beijing, 100083, China,
	Center of Materials Science and Optoelectronics Engineering $\&$ School of Microelectronics, Beijing Key Laboratory of Semiconductor Neural Network Intelligent Sensing and Computing Technology, Beijing 100083, China, University of Chinese Academy of Sciences, Beijing, 100049, China, and Cognitive Computing Technology Joint Laboratory, Wave Group, Beijing, 102208, China (E-mail: wangchangshuo@semi.ac.cn).}


\thanks{Baoli Lu and Liping Zhang are with the Institute of Semiconductors, Chinese Academy of Sciences, Beijing, 100083, China, 
	and Beijing Key Laboratory of Semiconductor Neural Network Intelligent Sensing and Computing Technology, Beijing, 100083, China (E-mail: lubaoli@semi.ac.cn; zliping@semi.ac.cn).}

\thanks{Xin Ning is with the Institute of Semiconductors, Chinese Academy of Sciences, Beijing, 100083, China, and Cognitive Computing Technology Joint Laboratory, Wave Group, Beijing, 102208, China (E-mail: ningxin@semi.ac.cn).}

\thanks{Xiao Bai is with the School of Computer Science and Engineering, Beihang University, Beijing, 100083, China (E-mail: baixiao@buaa.edu.cn).}

\thanks{\dag The author contributed equally to this work and should be considered co-first author.}

\thanks{*Shengwei Tian is the corresponding authors.}
}



\maketitle

\begin{abstract}
	In recent years, point cloud representation has become one of the research hotspots in the field of computer vision, and has been widely used in many fields, such as autonomous driving, virtual reality, robotics, etc. Although deep learning techniques have achieved great success in processing regular structured 2D grid image data, there are still great challenges in processing irregular, unstructured point cloud data. Point cloud classification is the basis of point cloud analysis, and many deep learning-based methods have been widely used in this task. Therefore, the purpose of this paper is to provide researchers in this field with the latest research progress and future trends. First, we introduce point cloud acquisition, characteristics, and challenges. Second, we review 3D data representations, storage formats, and commonly used datasets for point cloud classification. We then summarize deep learning-based methods for point cloud classification and complement recent research work. Next, we compare and analyze the performance of the main methods. Finally, we discuss some challenges and future directions for point cloud classification.
\end{abstract}


\section{Introduction}
\label{sec1}

In recent computer vision field, the processing technology of two-dimensional images is close to maturity \cite{zhang2021joint} \cite{zhou2020learning} \cite{ning2022hcfnn} \cite{wang2021brief}, and many researchers have shifted their research focus to three-dimensional scenes that are more in line with the real world. In a 3D scene, point cloud \cite{wangchanshuo20223d} plays an important role in representing the 3D scene because of its rich expression information. Therefore, point cloud has become a common form of data expression in the study of 3D vision. As technology advances, the acquisition of point cloud data is becoming increasingly
\begin{figure}[htbp]
	\centering
	\includegraphics[scale=0.25]{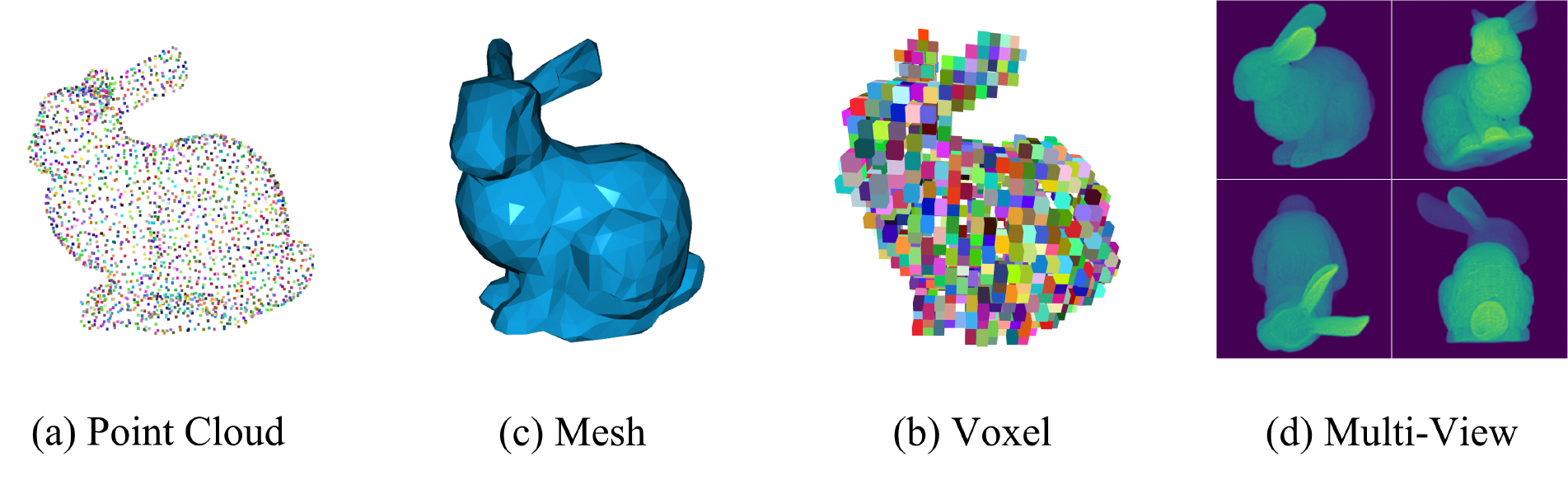}
	\caption{3D data representation}
	\label{f1}
\end{figure} 
intelligent and convenient, and there are many acquisition methods, such as: LIDAR laser detection, point cloud acquisition through 3D model calculation, point cloud acquisition through 3D reconstruction through 2D images \cite{cai2021voxel}, etc.

As the most basic point cloud analysis task, point cloud classification has been widely used in many fields such as security detection \cite{yan2021beyond} \cite{ning2021jwsaa}, target object detection \cite{9380436} \cite{zhou2018voxelnet}, medicine \cite{guo2018deep} \cite{yu20213d}, and three-dimensional reconstruction  \cite{xu2021review} \cite{yang2016automated}. The purpose of point cloud classification is to equip each point in the point cloud with a marker to identify the overall or part properties of the point cloud. Since the component attributes of point clouds belong to the category of point cloud segmentation, in this paper, we mainly focus on the overall attributes of point clouds, namely point cloud classification.

As shown in Fig.~\ref{f1}, 3D data comes in a variety of representations. Currently, it is possible to convert point clouds into mesh, voxel, or multi-view data to learn 3D object representation through indirect methods, but these methods are prone to problems such as loss of 3D geometric information of objects or excessive memory consumption. Before PointNet, due to the disorder and irregularity of point cloud, deep learning technology cannot directly process point cloud. Early point cloud processing used hand-designed rules for feature extraction, and then used machine learning-based classifiers (such as Support Vector Machines (SVM) \cite{hearst1998support}, AdaBoost \cite{kumar2012classification}, Random Forest (RF) \cite{gan2015random}, etc.) to predict the class label of the point cloud, but these Class methods have poor adaptive ability and are prone to noise \cite{zhou2022information}. Some researches solved the noise problem by synthesizing context information, such as Conditional Random Field (CRF) \cite{plath2009multi}, Markov Random Field (MRF) \cite{munoz2009contextual}, etc., which improved the classification performance to a certain extent. However, the feature expression ability of hand-designed rule extraction is limited, especially in complex scenes, the accuracy and generalization ability of the model cannot meet the requirements of human beings, and this method relies heavily on researchers with professional knowledge and experience.

With the rapid development of computer computing and data processing capabilities, the application of deep learning technology in point cloud analysis has also been promoted. The paper published by Charles et al. \cite{qi2017pointnet} of Stanford University in 2017 proposes a deep learning network, PointNet, that directly processes point clouds. This paper is a landmark, and methods for directly processing point clouds gradually dominate.

\begin{figure*}[!t]
\centering
\includegraphics[scale=0.5]{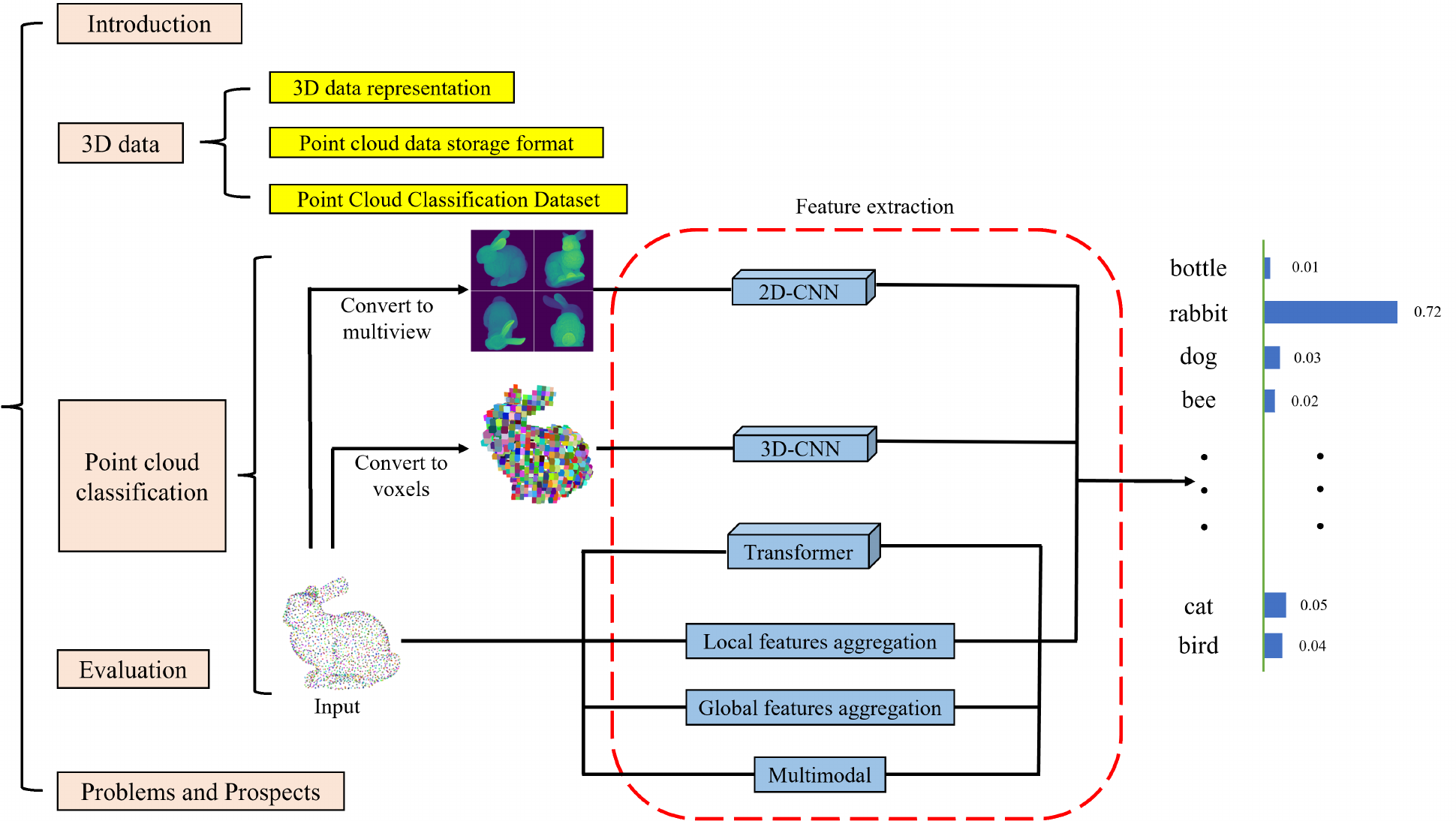}
\caption{Overall structure of the article. First, the point cloud is used as input, and the point cloud can be transformed by voxel or multi-view. Secondly, features are extracted from the original point cloud, the transformed voxels or multi-views. The final output is the probability value of each class.}
\label{f2}
\end{figure*}

Faced with the irregularity, disorder, and sparsity of 3D point clouds, point cloud classification is still a challenging problem. There are currently some reviews that analyze and summarize deep learning-based 3D point cloud classification methods. This paper improves on the previous work and adds new deep learning-based 3D point cloud classification methods, such as the recently popular transformer-based method. Finally, the future research direction of 3D point cloud classification technology is prospected. The overall structure of the article is shown in Fig.~\ref{f2}.

Specifically, the main contributions of our work are as follows:

$\bullet$ We first give a detailed introduction to the 3D data and make a deeper interpretation of the point cloud for the reader's understanding, and then give the datasets used for point cloud classification and their acquisition methods.

$\bullet$ We summarize recently published research on point cloud classification reviews, building on which to complement state-of-the-art research methods. These methods are classified into four categories according to their characteristics, including multi-view-based, voxel-based, point-cloud-based methods, and polymorphic fusion-based methods. And then the point-cloud-based methods are subdivided.

$\bullet$ We discuss the advantages and limitations of subcategories of methods based on their classification. This classification is more suitable for researchers to explore these methods on actual needs.

$\bullet$ We give the evaluation metrics and performance comparisons of the methods to better demonstrate the performance of various methods on the dataset, and then analyze some current challenges and future trends in this field.

\section{3D data}\label{sec2}

\subsection{3D data representation}\label{subsec2.1}

There are various representations of 3D data \cite{bai20223d}, such as point cloud, mesh, and voxel. Here we introduce them.

Point cloud:
A point cloud is essentially a large collection of tiny points drawn in 3D space, as shown in Fig.~\ref{f1}(a), which consists of a large collection of points captured using a 3D laser scanner. These points can express the spatial distribution and surface characteristics of the target. Each point in the point cloud contains rich information, such as: three-dimensional coordinates (x, y, z), color information (r, g, b) and surface normal vector, etc.

Mesh:
3D data can also be represented by a mesh grid, which can be viewed as a collection of points that build local relationships between points. Triangular mesh, also known as triangular patch (as shown in Fig.~\ref{f1}(b)), is one of the commonly used mesh grids to describe 3D objects. A collection of points and edges of a slice is called a mesh.

Voxel:
In 3D object representation, voxels are also an important form of 3D data representation, as shown in Fig.~\ref{f1}(c), voxels are good at representing non-uniformly filled regularly sampled spaces, therefore, voxels can effectively represent point cloud data with a lot of empty or evenly filled spaces. By voxelizing the point cloud data, it is beneficial to increase the data computing efficiency and reduce the access to random memory, but the voxelization of the point cloud data will inevitably bring a certain degree of information loss.

Multi-view:
A multi-view image (as shown in Fig.~\ref{f1}(d)) is also a representation of point cloud data, which is derived from a single-view image, and is an image that renders a 3D object into multiple viewpoints at a specific angle. The challenges are mainly the choice of perspective and perspective fusion.

\subsection{Point cloud data storage format}\label{subsec2.2}
There are hundreds of 3D file formats available for point clouds, and different scanners produce raw data in many formats. The biggest difference between point cloud data files is the use of ASCII and binary. Binary systems store data directly in binary code. Common point cloud binary formats include FLS, PCD, LAS, etc. Several other common file types can support both ASCII and binary formats. These include PLY, FBX. The E57 stores data in both binary and ASCII formats, combining many of the advantages of both in one file type. Below we introduce some commonly used point cloud data storage formats:

Obj: 
The point cloud file in obj format is developed by Wavefront Technologies. It is a text file. It is a simple data format that only represents the geometry, normal, color, and texture information of the 3D data. This format is usually represented in ASCII, but there are also proprietary obj binary versions.

Las: 
The las format is mainly used to store LIDAR point cloud data, which is essentially a binary format file. A LAS file consists of three parts: the header file area (including total number of points, data range, dimension information of each point), variable-length record area (including coordinate system, extra dimensions, etc.), and point set record area(including point coordinate information, R, G, B information, classification information, intensity information, etc.), the las format takes into account the characteristics of LIDAR data, the structure is reasonable, and it is easy to expand.

Ply: 
The full name of ply is Polygon File Format, which is inspired by obj and is specially used to store 3D data. Ply uses a nominally flat list of polygons to represent objects. It can store information including color, transparency, surface normal vector, texture coordinates and data confidence, and can set different properties for the front and back sides of the polygon. There are two versions of this file, an ASCII version, and a binary version.

E57: 
E57 is a vendor neutral file format for point cloud storage. It can also be used to store image and metadata information generated by laser scanners and other 3D imaging systems and is a strict format using fixed-size fields and records. It saves data using ASCII and binary codes and provides most of the accessibility of ASCII and the speed of binary, and it can store 3D point cloud data, attributes, images.

PCD: 
PCD is the official designated format of Point Cloud Library. It consists of two parts: header file and point cloud data. It is used to describe the overall information of point cloud. It has two data storage types, ASCII, and binary, but the header file of PCD file must use ASCII. Encoding, a nice benefit of PCD is that it adapts well to PCL, resulting in the highest performance compared to PCL applications.

\subsection{3D point cloud public datasets}\label{subsec2.3}

Today, there are many point cloud datasets provided by industries and universities. The performance of different methods on these datasets reflects the reliability and accuracy of the methods. These datasets consist of virtual or real scenes, which can provide ground-truth labels for training the network. In this section, we will introduce some commonly used point cloud classification datasets, and the division of each dataset is shown in Table.~\ref{tabl}.

ModelNet40 \cite{wu20153d}:  
The dataset was developed by the Vision and Robotics Laboratory at Princeton University. The ModelNet40 dataset consists of synthetic CAD objects. As the most widely used benchmark for point cloud analysis, ModelNet40 is popular for its diverse categories, clear shapes, and well-structured datasets. The dataset consists of objects of 40 categories (e.g. airplane, car, plant, lamp), of which 9843 are used for training and 2468 are used for testing. The corresponding points are uniformly sampled from the mesh surface and then further preprocessed by moving to the origin and scaling to a unit sphere.                                                                                    
Download link: \url{https://modelnet.cs.princeton.edu/}

ModelNet-C \cite{sun2022benchmarking}:
The ModelNet-C set contains 185,000 different point clouds and was created based on the ModelNet40 validation set. This dataset is mainly used to benchmark damage robustness for 3D point cloud recognition, with 15 damage types and 5 severity levels for each damage type, such as noise, density, etc. Helps to understand the robustness of the model.
Download link: \url{https://sites.google.com/umich.edu/modelnet40c}

ModelNet10 \cite{wu20153d}: 
ModelNet10: ModelNet10 is a subset of ModelNet40, the dataset contains only 10 classes and it is divided into 3991 training and 908 testing shapes.
Download link: \url{https://modelnet.cs.princeton.edu/}

Sydney Urban Objects \cite{de2013unsupervised}:  
The dataset, collected in Sydney CBD, contains a variety of common urban road objects, including 631 scanned objects in the categories of vehicles, pedestrians, signs, and trees.
Download link: \url{https://www.acfr.usyd.edu.au/papers/SydneyUrbanObjectsDataset.shtml}

ShapeNet \cite{chang2015shapenet}:  
ShapeNet is a large repository of 3D CAD models developed by researchers at Stanford University, Princeton University, and the Toyota Institute of Technology in Chicago, USA. The repository contains over 300 million models, of which 220,000 models are classified into 3,135 classes arranged using WordNet hypernym-hyponymy relations. ShapeNetCore is a subset of ShapeNet that includes nearly 51,300 unique 3D models. It provides 55 common object categories and annotations. ShapeNetSem is also a subset of ShapeNet, which contains 12,000 models. It is smaller in scale but has a wider coverage, including 270 categories.
Download link: \url{https://shapenet.org/}

ScanNet \cite{dai2017scannet}:  
ScanNet is an instance-level indoor RGB-D dataset containing 2D and 3D data. It is a collection of labeled voxels, not points or objects. As of now, ScanNet v2, the latest version of ScanNet, has collected 1513 annotated scans with about 90$\%$ surface coverage. In the semantic segmentation task, this dataset is labeled with 20 classes of annotated 3D voxelized objects.
Download link: \url{http://www.scan-net.org/}

ScanObjectNN \cite{uy-scanobjectnn-iccv19}:  
ScanObjectNN is a real-world dataset consisting of 2902 3D objects divided into 15 categories, which is a challenging point cloud classification dataset due to background, missing parts, and deformations in the dataset.
Download link: \url{https://hkust-vgd.github.io/scanobjectnn/}.

\begin{table}[htbp]
\centering
\caption{Point cloud classification datasets}
\resizebox{85mm}{!}{
	\begin{tabular}{cccccccc}
		\midrule
		Dataset & Year & Samples & Classes & Training & Test & Type  & Form \\
		\midrule
		ModelNet40\cite{wu20153d} & 2015  & 12311 & 40    & 9843  & 2468  & Synthetic & Mesh \\
		\midrule
		ModelNet40-C\cite{sun2022benchmarking} & 2015  & 185000 & 15    & - & - & Synthetic & Point Cloud \\
		\midrule
		ModelNet10\cite{wu20153d} & 2015  & 4899  & 10    & 3991  & 605   & Synthetic & Mesh \\
		\midrule
		SyDney Urban Objects\cite{de2013unsupervised} & 2013  & 588   & 14    & - & - & Real Word & Point Cloud \\
		\midrule
		ShapeNet\cite{chang2015shapenet} & 2015  & 51190 & 55    & - & - & Synthetic & Mesh \\
		\midrule
		ScanNet\cite{dai2017scannet} & 2017  & 12283 & 17    & 9677  & 2606  & Real Word & RGB-D \\
		\midrule
		ScanObjectNN\cite{uy-scanobjectnn-iccv19} & 2019  & 2902  & 15    & 2321  & 581   & Real Word & Point Cloud \\
		\bottomrule
\end{tabular}}%
\label{tabl}%
\end{table}%

\section{Point cloud classification method based on deep learning}\label{sec3}

\begin{figure*}[h]
\centering
\includegraphics[scale=0.5]{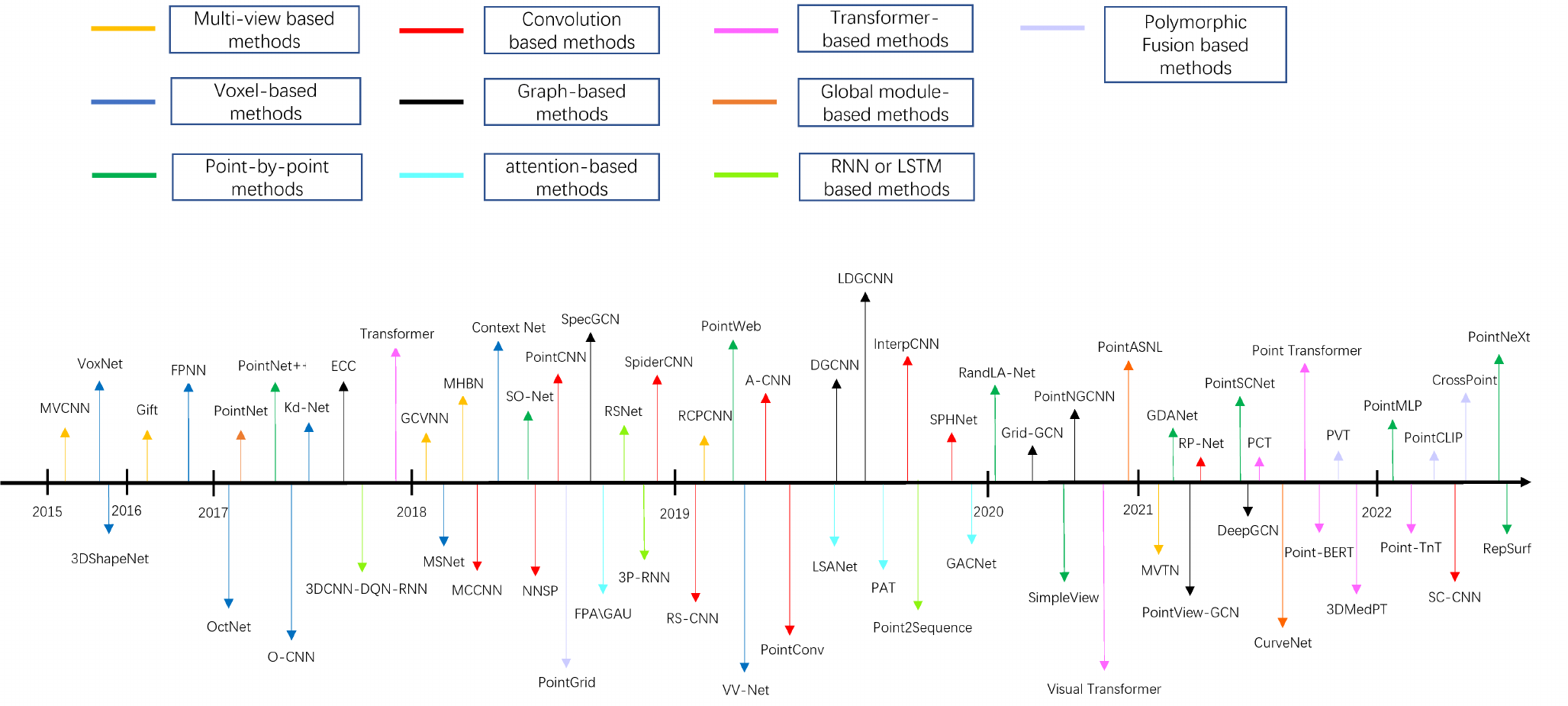}
\caption{Timeline of the development of classification methods}
\label{f3}
\end{figure*}

The point cloud classification model based on deep learning \cite{wang2022uncertainty} has been widely used in point cloud analysis due to its advantages of strong generalization ability and high classification accuracy. This section provides a detailed division of deep learning-based point cloud classification methods and supplements some recent research works. Fig.~\ref{f3} shows the publication timeline of each classification method.

\subsection{Multi-view-based methods}
\begin{figure}[h]
\centering
\includegraphics[scale=0.26]{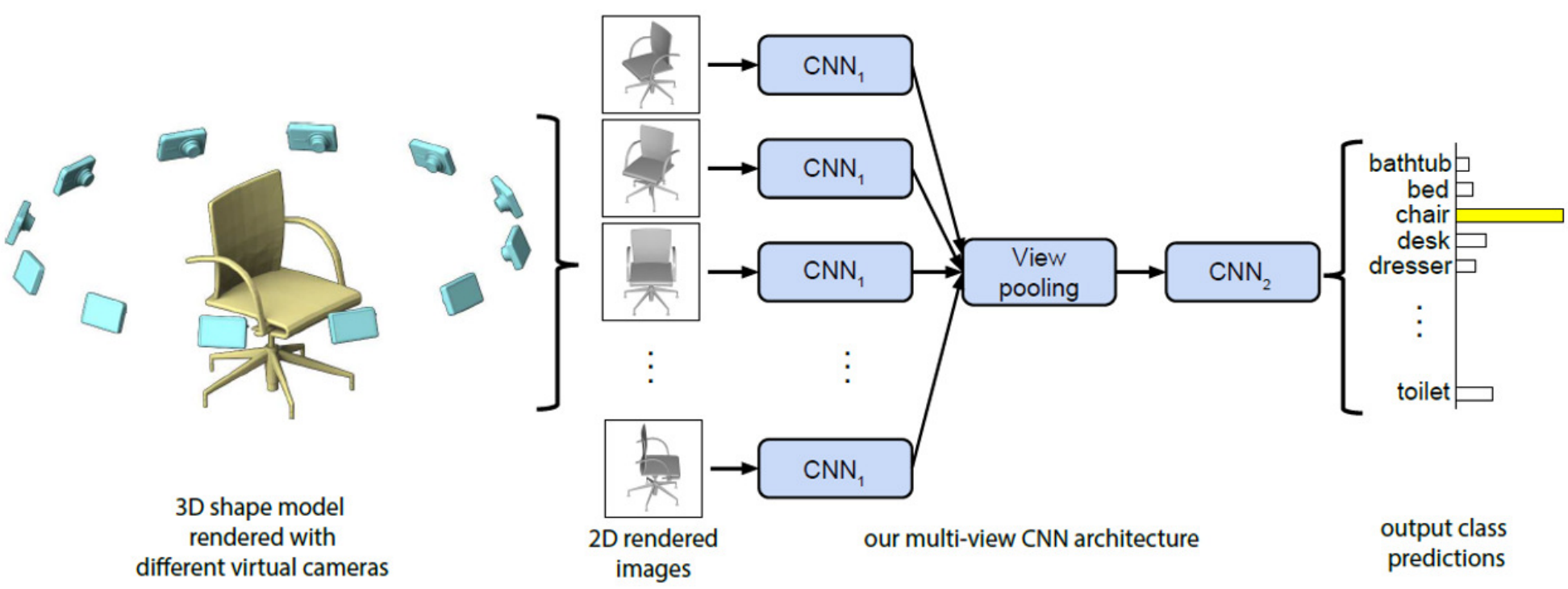}
\caption{Schematic diagram of MVCNN\cite{2015Multi} structure}
\label{f4}
\end{figure}

Multi-view learning is a machine learning framework in which data is represented by multiple distinct feature groups, each of which is called a specific view. The multi-view-based method is a deep learning based on 2D images. This method is divided into three steps: First, project the 3D image into multiple views. Second, extract the view function. Third, fuse these functions to accurately classify 3D shapes.
In 2015, Su et al. \cite{2015Multi} first proposed the multi-view convolutional neural network MVCNN method. Since the collection of 2D views can provide a lot of information for 3D shape recognition, this method recognizes 3D shapes from the collection of rendering views on 2D images. Representing 3D Shapes" is a longstanding problem. This method first demonstrates a standard CNN architecture trained to recognize shapes independently of rendered views and shows that 3D shapes can be recognized even from a single view. When multiple views of an object are provided, the recognition rate is further improved by using the CNN architecture to combine the information of multiple views into a compact shape descriptor. The network architecture is shown in Fig.~\ref{f4}. This method requires a large amount of computation in the projection retrieval process, and when converting multiple view features into global features through max-pooling, other non-maximum element information is ignored, so it will inevitably cause information missing.

Therefore, in view of the large amount of computation and efficiency of MVCNN, Bai et al. \cite{2016GIFT} proposed a real-time 3D shape search engine, namely GIFT, this method uses GPU acceleration in the stages of projection and view feature extraction, which greatly shortens the time spent on retrieval tasks, and has high efficiency and ability to handle large-scale data.

Based on MVCNN, the GVCNN architecture proposed by Feng et al. \cite{2018GVCNN} introduces the "view-group-shape" architecture to extract descriptors, which can effectively utilize the feature relationship between views. In order to overcome the problem of information loss due to max pooling, Wang et al. \cite{wang2019dominant} introduced the view clustering and pooling layer based on the dominant set to improve the MVCNN method and proposed RCPCNN. It entails building a view similarity graph, clustering nodes (views) in this graph based on dominance sets, and pooling information from within each cluster. The recurrent clustering and pooling layer are to aggregate multi-view features in a way that provides more discriminative power for 3D object recognition. This recurrent clustering and pooling convolutional neural network (RCPCNN) module is plugged into ready-made pre-trained CNN which improves the performance of multi-view 3D object recognition.

\begin{figure}[htbp]
\centering
\includegraphics[scale=0.26]{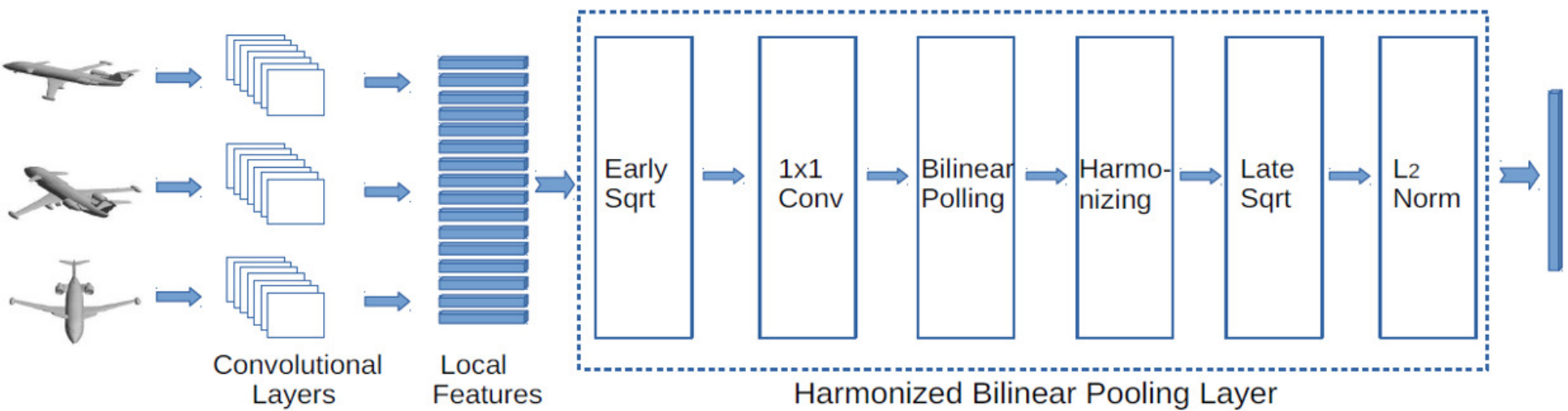}
\caption{Architecture of MHBN\cite{yu2018multi}}
\label{f5}
\end{figure}

The MHBN method proposed by Yu et al. \cite{yu2018multi} aggregates local convolutional features by bilinear pooling to obtain efficient 3D object representations. Fig.~\ref{f5} shows the network architecture of this method. To achieve an end-to-end trainable framework, this method coordinates the merging of bilinear pooling into one layer in the network, from the perspective of patch-to-patch similarity measurement, to address the problem of view-based methods pooling view features that differ.

To avoid the lack of dynamism of current multi-view methods, Hamdi et al. \cite{hamdi2021mvtn} propose the Multi-View Transition Network (MVTN), which proposes a differentiable module that predicts the best viewpoint for a task-specific multi-view network, regressing the best for 3D shape recognition Viewpoint. MVTN can be trained end-to-end with any multi-view network, showing significant performance gains in 3D shape classification and retrieval tasks without additional training supervision.

Summary: Compared with the traditional manual extraction feature classification, the multi-view-based method has a better effect in point cloud classification, but it is still difficult to make full use of information. The application of large-scale scenes, and the inherent geometric relationship of 3D data is the challenge we need to face.

\subsection{Voxel-based methods}

This type of approach converts a 3D point cloud model into voxel form that approximates the shape of an object, each voxel block contains a set of associated points, and then uses 3D CNNs to classify the voxels.

\begin{figure}[htbp]
\centering
\includegraphics[scale=0.32]{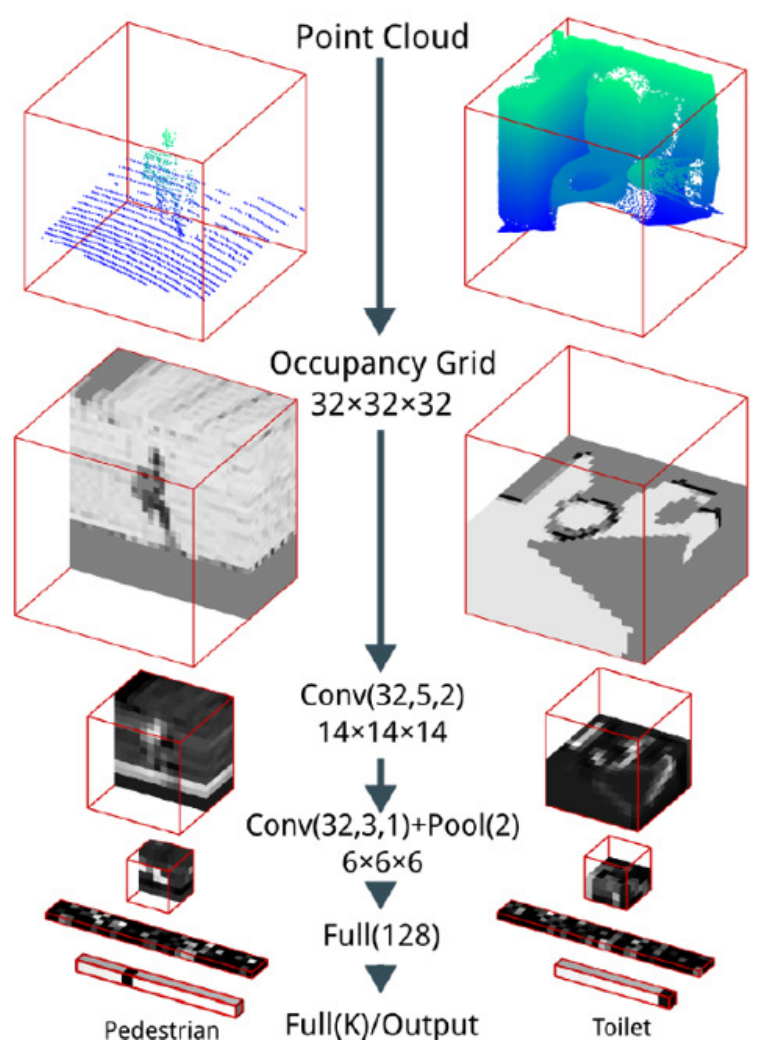}
\caption{Architecture of VoxNet\cite{maturana2015voxnet}}
\label{f6}
\end{figure}

Maturana et al. \cite{maturana2015voxnet} proposed a convolutional neural network architecture called VoxNet to represent 3D information with a volumetric occupancy grid. VoxNet is the earliest voxel-based 3DCNN model. As shown in Fig.~\ref{f6}, this method normalizes each grid, then constructs a feature map through convolution and max pooling a single voxel block. This architecture uses 2.5D to represent the features of the local description scan, and adopts a full volume representation, which improves the ability to express environmental information and enables powerful 3D object recognition.

Wu et al. \cite{wu20153d} proposed 3D ShapeNets to recognize 3D objects. The model represents a 3D shape as a probability distribution of binary variables on a grid of 3D voxels, each of which can be represented by a binary tensor, and predicts the next best view in the presence of uncertain initial recognition. Finally, 3D ShapeNets can incorporate new views to identify objects in common with all views.

Since 3D convolutions are computationally expensive, the spatial resolution increases model complexity when convolving 3D voxels. In response to this phenomenon, Li et al. \cite{li2016fpnn} proposed a field detection neural network (FPNN), which represented 3D data as a 3D field, then sampled the input field through a set of detection filters, and finally used the field detection filter to characterize extract. The field detection layer can be used with other inference layers, but this approach makes the semantic segmentation results lower resolution.

To reduce memory consumption and improve computational efficiency, some scholars use octree structure instead of fixed-resolution voxel structure. Riegler et al. \cite{riegler2017octnet} proposed OctNet, which exploits the sparsity of 3D data to hierarchically partition the space with a set of unbalanced octrees, where each leaf node stores a pooled feature representation. This method enables deeper networks without affecting their resolution. Wang et al. \cite{wang2017cnn} proposed an octree-based convolutional neural network O-CNN, which uses octrees to represent 3D data information and discretize its surface. The 3D CNN operation is only performed on the octant occupied by the 3D shape surface, which improves the computational efficiency and power.

Following the use of the octree structure in the 3D data representation, the Kd-tree structure is also used in the point cloud classification model. The Kd-network proposed by Klokov et al. \cite{klokov2017escape} adopts, compared with voxel and mesh, the ability of Kd-tree to index and structure 3D data has been improved, so Kd-network has a smaller memory footprint and more efficient computation during training and testing. The 3D Context Net proposed by Zeng et al. \cite{zeng20183dcontextnet} utilizes the method of local and global context cues imposed by Kd-tree for semantic segmentation.

Octree structure and Kd-tree structure reduce memory consumption to a certain extent and improve computational efficiency, but due to the influence of voxel boundary value, these two structures cannot make full use of local data features and the accuracy is affected . Wang et al. \cite{wang2018msnet} proposed a multi-scale convolutional network, MSNet. This method first divides the space into voxels of different scales, then applies MSNet on multiple scales simultaneously to learn local features, and finally uses a conditional random field (CRF) The prediction results of MSNet are globally optimized to achieve a more accurate point cloud classification task.

The VV-Net proposed by Meng et al. \cite{meng2019vv} uses a kernel-based interpolating variational autoencoder (VAE) to encode localities in voxels, each voxel is further subdivided into sub-voxels, which are within voxels interpolate sparse point samples to efficiently handle noisy point cloud datasets.

Summary: Compared with the multi-view method, the voxel-based method pays attention to the relationship between 3D data, and groups the point clouds with internal connections into a set of points, thereby establishing voxels. Although the voxel-based model solves the point cloud disorder and unstructured problem, the sparseness and incomplete information of point cloud data lead to the low efficiency of the classification task, so the information in the point cloud cannot be fully utilized.

\subsection{Point cloud-based method}

Many current research methods focus more on directly processing point clouds using deep learning techniques. Feature aggregation operator is the core of processing point cloud, which realizes the information transfer of discrete points. Feature aggregation operators are mainly divided into two categories: local feature aggregation and global feature aggregation. In this section, from the perspective of feature aggregation, the two categories of methods are divided and introduced in more detail.

\begin{figure}[htbp]
\centering
\includegraphics[scale=0.26]{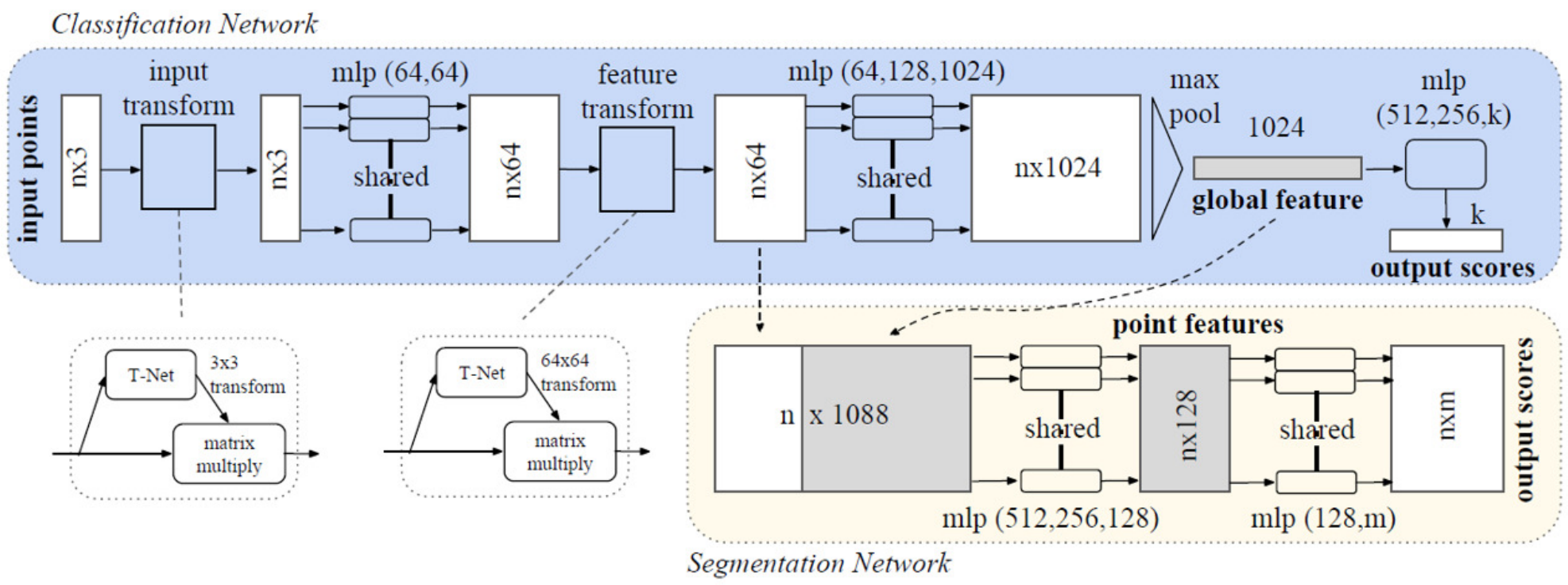}
\caption{PointNet\cite{qi2017pointnet} network architecture}
\label{f7}
\end{figure}

In 2017, the PointNet proposed by Qi et al. \cite{qi2017pointnet} (shown in Fig.~\ref{f7}) is a pioneering study of point cloud-based methods, which is a method of global feature aggregation. This method directly takes the point cloud as input, transforms it through the T-Net module, then learns each point by sharing the full connection, and finally aggregates the features of the point into global features through the maximum pooling function. Although PointNet is a pioneer based on deep learning, it still has defects. For example, PointNet only captures the feature information of a single point and global points but does not consider the relational representation of adjacent points, which makes PointNet unable to effectively perform fine-grained classification.

\subsubsection{Local feature aggregation}
\paragraph
1 Point-by-point method

\begin{figure}[htbp]
\centering
\includegraphics[scale=0.26]{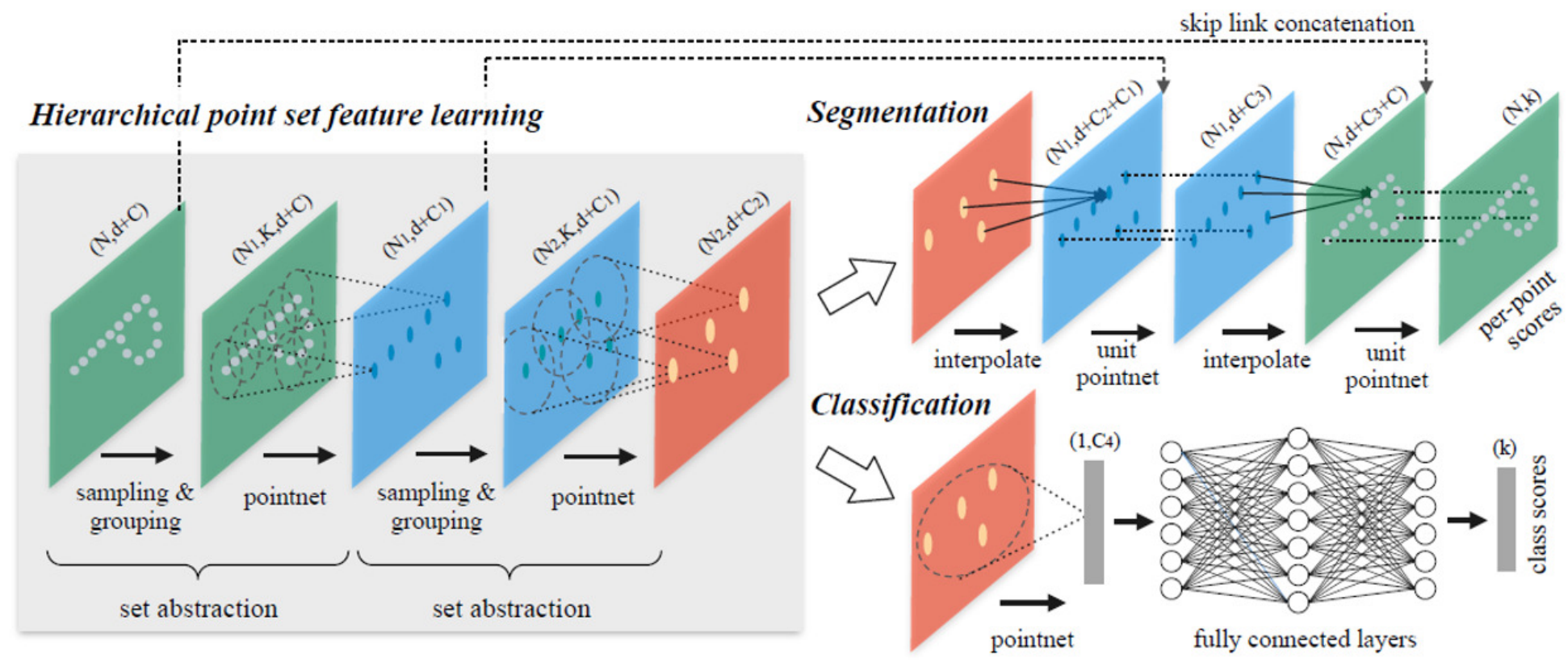}
\caption{PointNet++\cite{qi2017pointnet++} point cloud segmentation and classification architecture diagram}
\label{f8}
\end{figure}

Qi et al.\cite{qi2017pointnet++} successively proposed PointNet++ based on PointNet. This method processes point clouds in a hierarchical manner, and each layer consists of a sampling layer, a grouping layer, and a PointNet layer. Among them, the sampling layer obtains the centroid of the local neighborhood, the grouping layer constructs a subset of the local neighborhood, and the PointNet layer obtains the relationship between the points in the local area, as shown in Fig.~\ref{f8}. PointNet++ needs to solve two problems: dividing the generated point set and aggregating local features through a local feature learner. But PointNet++ still does not ignore the prior relationship between points.

\begin{figure}[htbp]
\centering
\includegraphics[scale=0.26]{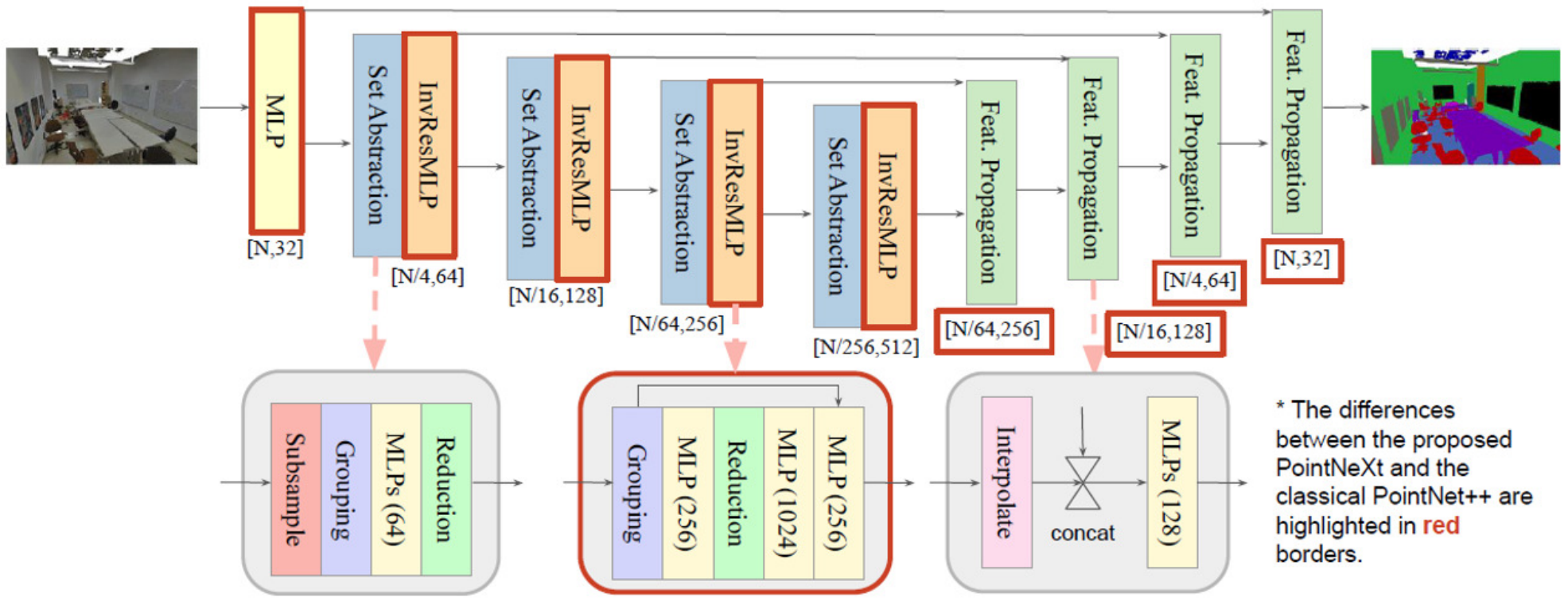}
\caption{PointNeXt\cite{qian2022pointnext} architecture}
\label{f9}
\end{figure}

Based on PointNet++, Qian et al.\cite{qian2022pointnext} improved the training and training strategy to improve the performance of PointNet++ and introduced a separable MLP and an inverted residual bottleneck design in the PointNet++ framework, and named its framework PointNeXt (as shown in Fig.~\ref{f9}).

Zhao et al.\cite{zhao2019pointweb} proposed a method of extracting features from the local neighborhood of point clouds - PointWeb, which connects the points in the local neighborhood with each other, and proposes a new module, Adaptive Feature Adjustment (AFA), which is used to find information transfer between points. This method makes full use of the local features of points and realizes adaptive adjustment.

Hu et al. \cite{hu2020randla} proposed an efficient and lightweight neural architecture - RandLA-Net, which uses random point sampling, and increases the receptive field of each point through an efficient local feature aggregation module, so as to better capturing complex local structures reduces memory footprint and computational cost, but this approach may discard some key features of sparse points.

The SO-Net proposed by Li et al. \cite{li2018so} constructs a self-organizing map (SOM), extracts the features of each point and SOM node hierarchically, and uses a single feature vector to represent the point cloud, by appending a 3-layer perception to the encoded global feature vector Multi-Layer Perceptron(MLP) is used to classify point clouds. SO-Net has good parallelism and simple structure, but it has limitations in processing large-scale point cloud data.

In order to fully capture the most critical geometric information, Xu et al. \cite{xu2021learning} proposed the geometric disentanglement attention network GDANet, and introduced the Geometry-Disentangle module to decompose the original point cloud into two parts: contour and plane, so as to capture and refine 3D semantics to supplementing local information. The method has good robustness. 

Goyal et al. \cite{goyal2020revisiting} proposed a projection-based method, SimpleView, which showed that training and evaluation factors independent of the network architecture have a large impact on point cloud classification performance. The PointSCNet proposed by Chen et al. \cite{chen2021pointscnet} is used to capture the geometric information and local area information of the point cloud. It consists of three modules: a space-filling curve-guided sampling module, an information fusion module, and a channel spatial attention module. In PointSCNet the raw point cloud is fed to the sampling and grouping block, which is sampled using a Z-order curve to obtain the high correlation between points and local regions. After extracting the sampled point cloud features, a feature fusion module is designed to learn the structure and related information. Finally, the keypoint features are enhanced by the channel space module.

Ma et al. \cite{ma2022rethinking} noticed that detailed local geometric information may not be the key to analyzing point clouds, so they introduced a pure residual network called PointMLP, which does not have a complex local geometry extractor, but is equipped with a lightweight geometric affine module, there is a significant improvement in inference speed. Huang et al. \cite{huang2022shape} calculated the rate of change of recognition confidence when a shape-invariant perturbation with explicit constraints was added to each point in the point cloud, and according to this method, they proposed a point cloud sensitivity map, which was then used to propose a shape-invariant adversarial point cloud.

Ran et al. \cite{ran2022surface} proposed using RepSurf (representative surfaces) to represent point clouds, which has two variants: Triangular RepSurf and Umbrella RepSurf. This representation can be used as a module in the point cloud classification and segmentation framework to improve the performance of the point cloud framework.

\paragraph
2 Convolution-based methods

Convolutional Neural Network (CNN) plays an important role in deep learning and is the most basic deep learning model. Its excellent performance in the field of 2D image processing has led researchers to apply it to 3D point clouds and design point convolution for point cloud classification.

Atzmon et al. \cite{atzmon2018point} proposed the point convolutional neural network (PCNN), which applied convolutional Neural Network (CNN) to point cloud. First, the function on the point cloud is extended to a continuous volume function in space; then a continuous volume convolution is applied to the function; the final result is a constrained point cloud, as shown in Figure 13.

Liu et al. \cite{liu2019relation} proposed Relational Shape Convolutional Networks (RS-CNN) to extend regular CNNs to irregular point clouds for analysis of point clouds. Yousefhussien et al. \cite{yousefhussien2018multi} proposed an one-dimensional fully convolutional network. Wang et al. \cite{wang2018deep} proposed a deep neural network DNNSP with spatial pooling to classify large-scale point clouds. This method can learn the features from the entire region to the center point of the point cluster, to achieve a robust representation of point features. Komarichev et al. \cite{komarichev2019cnn} propose a point cloud-based annular convolutional neural network (A-CNN) model. Ran et al. \cite{ran2021learning} proposed RPNet based on a group relation aggregation module, which is robust to rigid transformations and noise. The SCN (ShapeContextNet) proposed by Xie et al. \cite{xie2018attentional} is represented by using the shape context as a building block, so that it can capture and propagate object part information. SCN is an end-to-end model.

Since directly convolving kernels with point-related features would result in the discarding of shape information and variance in point ordering, Li et al. \cite{li2018pointcnn}  proposed PointCNN to address this problem, which confirmed the development of local structures for point cloud classification networks importance. 

Due to the sparsity, irregularity, and disorder of the point cloud, it is difficult to directly perform the convolution operation on it. Wu et al. \cite{wu2019pointconv} proposed to apply the dynamic filter to the convolution operation called PointConv, which is simple and reduces the computer storage pressure. MCCNN \cite{hermosilla2018monte} represents the convolution kernel itself as a multilayer perceptron and describes the convolution as a Monte Carlo integration. SpiderCNN \cite{xu2018spidercnn} inherits the multi-scale hierarchical structure of CNN and consists of SpiderConv units, which extend the convolution operation from regular grids to irregular point sets that can be embedded in n-dimensional space by parameterizing a series of convolution filters to effectively extract geometric features from point clouds. Mao et al. \cite{mao2019interpolated} designed interpolating convolutional neural networks (InterpCNNs) based on combining InterConv(interpolating convolution operation). 

Esteves et al. \cite{esteves2018learning} use multi-valued spherical functions to model 3D data, and propose a spherical convolutional network that implements them by implementing precise convolutions on spheres in the spherical harmonic domain, thus solving the 3D rotation variance problem in convolutional neural networks. SPHNet \cite{poulenard2019effective} is based on PCNN to achieve rotation invariance by using spherical harmonics in different layers of the network. 

Since the local features of point clouds are difficult to aggregate and transfer effectively, Wang et al.\cite{wang2022learning} proposed a spatial coverage convolutional neural network (SC-CNN), the core of which is the spatial coverage convolution (SC-Conv). Anisotropic spatial geometry is constructed in the point cloud to implement depthwise separable convolution, replacing depthwise convolution with the spatial coverage operator (SCOP).

\paragraph
3 Graph-based methods

The graph neural network (GNN) was first proposed by Scarselli et al. \cite{scarselli2008graph}. Bruna et al. \cite{bruna2013spectral} are the first to apply convolution to low-dimensional graph structures to effectively represent depth. Kipf et al. \cite{kipf2016semi} further proposed that a graph convolutional network (GCN) works well in semi-supervised classification tasks, and in fact, GCN is an optimization of CNN. 

Simonovsky \cite{simonovsky2017dynamic} proposed an ECC(edge conditional convolutional) network that can be applied to any graph structure in combination with the application of edge labels. This method uses the points as the vertices of the graph and the distance between the points as the weight, and performs a weighted average convolution operation, using the maximum sampling the information of aggregated vertices. It can be used for large-scale point cloud segmentation, but the amount of computation is large.

SpecGCN \cite{wang2018local} replaces the standard max-pooling step with a recursive clustering and pooling strategy. Grid-GCN \cite{xu2020grid}  uses a coverage-aware network query (CAGQ), which improves spatial coverage and reduces theoretical time complexity by exploiting the efficiency of grid space.

Mohammadi et al. \cite{mohammadi2021pointview} proposed PointView-GCN with a multi-level graph convolutional network (GCN) to hierarchically aggregate shape features of single-view point clouds, which facilitates encoding of object geometric cues and multi-view relationships, resulting in more specific global features.

Wang et al. \cite{wang2019dynamic} proposed a dynamic graph CNN (DGCNN) for point cloud learning and proposed an edge convolutional (EdgeConv) network module, which can better capture the local geometric features of point clouds and maintain arrangement invariance, which demonstrates the importance of local geometric features for 3D recognition tasks. Zhang et al. \cite{zhang2019linked} further optimized DGCNN and proposed a Linked Dynamic Graph Convolutional Neural Network (LDGCNN), which removed the transformation network in DGCNN to simplify the network model and optimized the network by connecting hierarchical features of different dynamic graphs, which can better solve the gradient vanishing problem. 

The PointNGCNN proposed by Lu et al. \cite{lu2020pointngcnn} describes the relationship between points in the neighborhood in a neighborhood graph and uses neighborhood graph filters to extract neighborhood feature information and spatial distribution information in feature space and Cartesian space.

\paragraph
4 Attention-based methods

The basic idea of the attention mechanism is to apply human perception to the machine, but human perception selectively focuses on part of the scene instead of processing the entire scene at a time, so researchers focus on the attention mechanism to carry out research and apply to the field of point cloud classification. 

Yang et al. \cite{yang2019modeling} developed a point attention transformer (PAT) based on point cloud reasoning. It is proposed to use efficient GSA (Group-Shuffle Attention) instead of expensive MHA (Multi-Head Attention) for modeling the relationship between points, and propose a method called Gumbel Subset Sampling (GSS) to select a subset of representative points, which reduces the computational cost.

Li et al. \cite{li2018pyramid} proposed Feature Pyramid Attention Module (FPA) and Global Attention Upsampling Module (GAU) by combining attention mechanism and spatial pyramid. 
Chen et al. \cite{chen2019lsanet} designed a Local Spatial Awareness (LSA) layer and proposed an LSANet network architecture based on the LSA layer. LSA can learn the spatial relationship layer in the local area to generate spatially distributed weights, so that spatially independent operations can be performed. The spatial information extraction function of this method is powerful. 

Wang et al. \cite{wang2019graph} proposed GACNet based on graph attention convolution (GAC). The GAPointNet proposed by Chen et al. \cite{chen2021gapointnet} combines the self-attention mechanism with graph convolution, learns local information representation by embedding a graph attention mechanism in stacked MLP layers, and uses a parallel mechanism to aggregate the attention features of different GAPLayer layers, where GAPLayer layers and attention layers can be embedded into existing trained models to better extract local contextual features from unordered point clouds.

\subsubsection{Global feature aggregation}

\paragraph
1 Transformer-based methods

Since the transformer was first proposed in 2017 \cite{vaswani2017attention}, it has achieved world-renowned results in the field of computer vision. Many researchers also use this structure in point cloud processing.

\begin{figure}[htbp]
\centering
\includegraphics[scale=0.26]{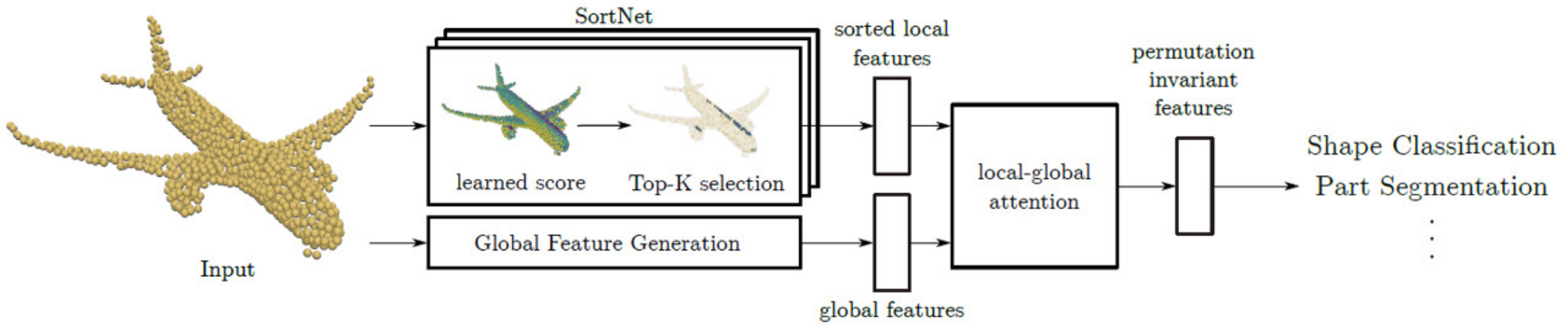}
\caption{Architecture of Point Transformer\cite{engel2021point}}
\label{f10}
\end{figure}

Engel et al. \cite{engel2021point} propose a deep neural network Point Transformer that operates directly on unordered and unstructured point sets, and propose a learning score-based focus module, ScorNet, as part of the Point Transformer. The point cloud is used as the input of the Point Transformer, and local and global features are extracted from it, and then SortNet is used to rank the local features, and finally the local global attention is used to associate the local global features, as shown in Fig.~\ref{f10}.

Berg et al. \cite{berg2022points} found that the self-attention operator grows rapidly and inefficiently with the growth of the input point set, and the attention mechanism is difficult to find the relationship between each point in the global, so they propose a two-stage method - Point TnT, this method enables a single point and a point set to pay attention to each other effectively.

The Visual Transformer (VT) proposed by Wu et al. \cite{wu2020visual}, which applies Transformer to label-based images from feature maps, can learn and associate high-level concepts of sparse distributions more efficiently. Carion et al. \cite{carion2020end} propose a method to treat object detection as a direct ensemble prediction problem called Detection Transformer (DETR), which is an end-to-end detection transformer that takes CNN features as input and uses a Transformer encoder-decoder to generate boundaries. 

Guo et al. \cite{guo2021pct} proposed a Transformer-based point cloud learning framework - Point Cloud Transformer (PCT), and proposed offset attention with implicit Laplacian operator and normalization refinement, the framework has permutation invariance and is more suitable for point cloud learning

3DMedPT is a Transformer network proposed by Yu et al. \cite{yu20213d} for 3D medical point cloud analysis. 

Inspired by BERT, Yu et al. \cite{yu2022point} proposed a new method for learning Transformer called Point-BERT. This method first divides the point cloud into several local blocks and generates discrete point labels of local information through a point cloud marker, and then by randomly masking some input point clouds and feeding them into the backbone Transformer, this method can generalize the concept of BERT to point clouds. 

Pang et al. \cite{pang2022masked} proposed Point-MAE, which is a masked autoencoder method for point cloud self-supervised learning to solve the problems caused by point cloud location information leakage and uneven information density. question. 

He et al. \cite{he2022voxel} introduced a voxel-based set attention module (VSA) to establish the Voxel Set Transformer (VoxSeT) architecture. VoxSeT can manage point clusters through the VSA module and process them in parallel with linear complexity. This method combines the high performance of Transformer with the high efficiency of voxel-based model, it has good performance in point cloud modeling.

\paragraph
2 Global module-based methods

Wang et al. \cite{wang2018non} propose a global module that computes the response at a location as a weighted sum of all location features, providing a solution to aggregating global features, but the global point-to-point mapping may still be insufficient to extract the underlying patterns implied by the point cloud shape.

\begin{figure}[htbp]
\centering
\includegraphics[scale=0.26]{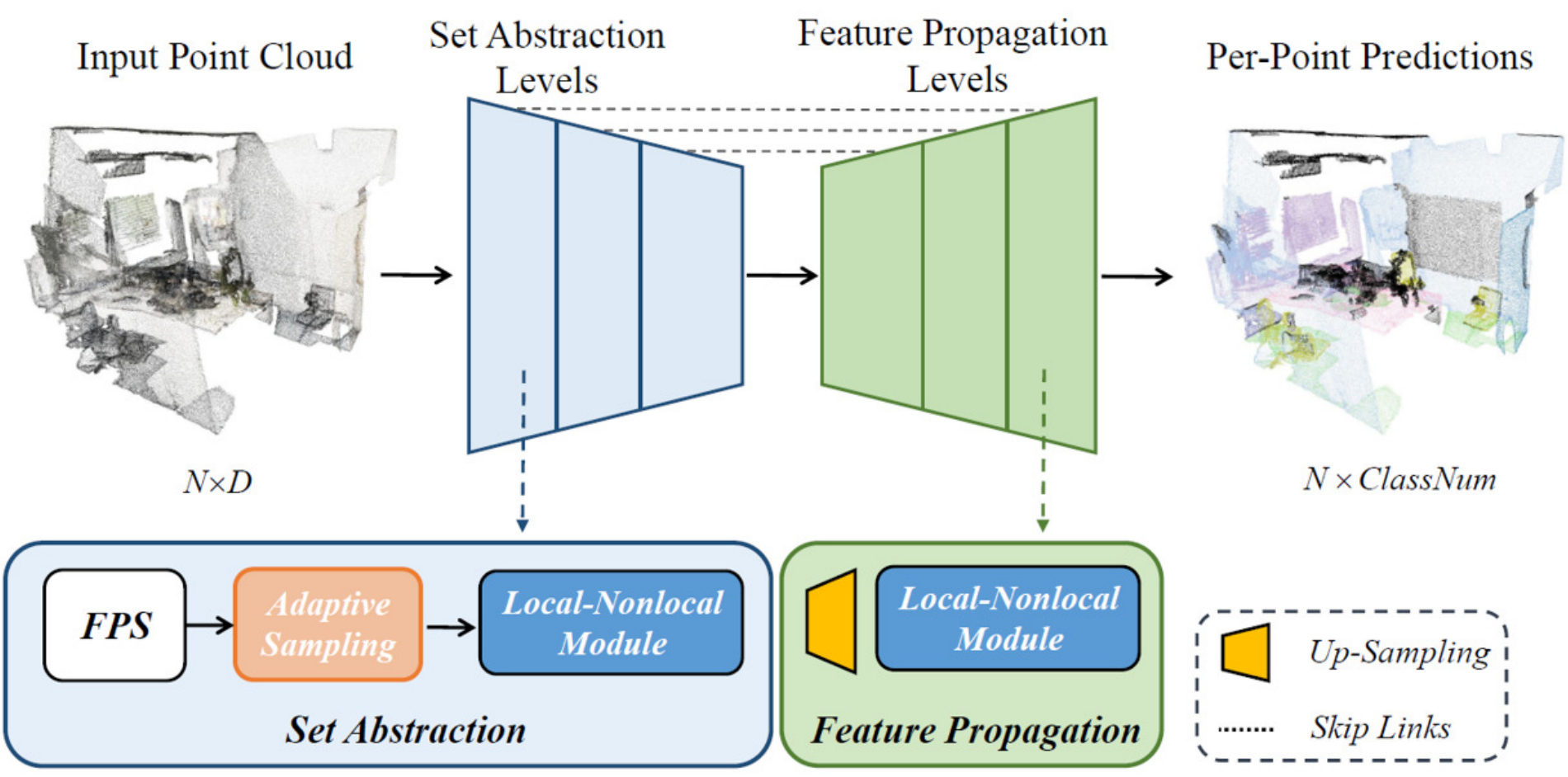}
\caption{Architecture of PointASNL\cite{yan2020pointasnl}}
\label{f11}
\end{figure}

Yan et al. \cite{yan2020pointasnl} propose an end-to-end network PointASNL that combines an adaptive sampling module (AS) and a local non-local module (L-NL), which can effectively deal with noisy point clouds. The AS module updates the features of the points by reasoning, then normalizes the weight parameters and re-weights the initial sampling points, which can effectively alleviate the bias effect. The L-NL module consists of local and non-local units of points, reducing the sensitivity of the learning process to noise. The PointASNL architecture is shown in Fig.~\ref{f11}.
Li et al. \cite{li2021deepgcns} adopted some CNN methods to support a deep GCN architecture, called DeepGCN, which consists of three blocks: GCN Backbone block for input point cloud feature transformation, fusion block for generating and fusing global features, MLP block prediction block used to predict labels. To solve the problem of gradient disappearance during GCN training, it is possible to train deeper GCN networks.

Xiang et al. \cite{xiang2021walk} proposed a method based on aggregating hypothetical curves in point clouds, CurveNet, and effectively implemented the aggregation strategy, including a curve grouping operator and a curve aggregation operator. The network consists of a bunch of building blocks, FPS represents the farthest point sampling method.

\paragraph
3 RNN or LSTM-based methods

RNN (Recurrent Neural Network) can usually effectively utilize context information to process sequence data. LSTM (Long Short-term Memory) is a special RNN, which can effectively solve the problems of gradient disappearance and gradient explosion in the training process of longer sequence data.

Engelmann et al. \cite{engelmann2017exploring} extended the input-level context information and output-level context information based on PointNet, which enables PointNet to be applied in large-scale scenarios. It can make PointNet applied in large-scale scenarios.

Liu et al. \cite{liu20173dcnn} proposed a 3DCNN-DQN-RNN method, which fuses 3D Convolutional Neural Network (CNN), Deep Q Network (DQN) and residual Recurrent Neural Network (RNN). First, the feature representation of points is obtained by 3DCNN. Second, DQN can detect and localize objects, and can automatically perceive the scene and adjust the feedback of 3DCNN. Finally, the RNN is used to identify the connections and differences of multi-scale features. Where LSTM (Long Short-term Memory) units are used to prevent vanishing gradients and make the network have long-term memory, the method improves the accuracy of processing large-scale point clouds.

The RSNet network proposed by Huang et al. \cite{huang2018recurrent} takes the original point cloud as input, and then performs feature extraction, and then passes through the slice pooling layers in the three directions of x, y, and z. Each layer uses a bidirectional RNN to extract local features, and then uses slice parsing. The layer assigns the features of the point cloud sequence to each point, and finally outputs the predicted semantic label of each point.

Ye et al. \cite{ye20183d} proposed an end-to-end semantic segmentation method, 3P-RNN, by combining CNN and RNN, which consists of a point pyramid module and a bidirectional hierarchical RNN module. In the work of distinguishing the same semantics, this method has certain limitations. 

Point2Sequence proposed by Liu et al. \cite{liu2019point2sequence} uses RNN to capture fine-grained contextual information to learn 3D shape features, it introduces an attention mechanism to enhance feature extraction.

\subsection{Polymorphic Fusion-based methods}

The strategy of combining PointGrid with grids proposed by Le et al. \cite{le2018pointgrid} is to mix points and grids for representation. PointGrid is composed of several convolution blocks, which represent the features of different layers through maximum pooling. Each convolution layer includes a convolution kernel, and the over-fitting phenomenon is controlled by the pooling layer, and then completed by full connection. For inference, PointGrid has two fully connected layers, and finally performs regression with a softmax classifier, which can better identify fine-grained models and represent local shapes.

Zhang et al. \cite{zhang2021pvt} proposed a novel point cloud learning method, PVT (Point-Voxel Transformer), combining sparse window attention module (SWA) and relative attention module (RA), which combines voxel-based and point-based model idea, this method excels in the accuracy of point cloud classification. 

\begin{figure}[htbp]
\centering
\includegraphics[scale=0.26]{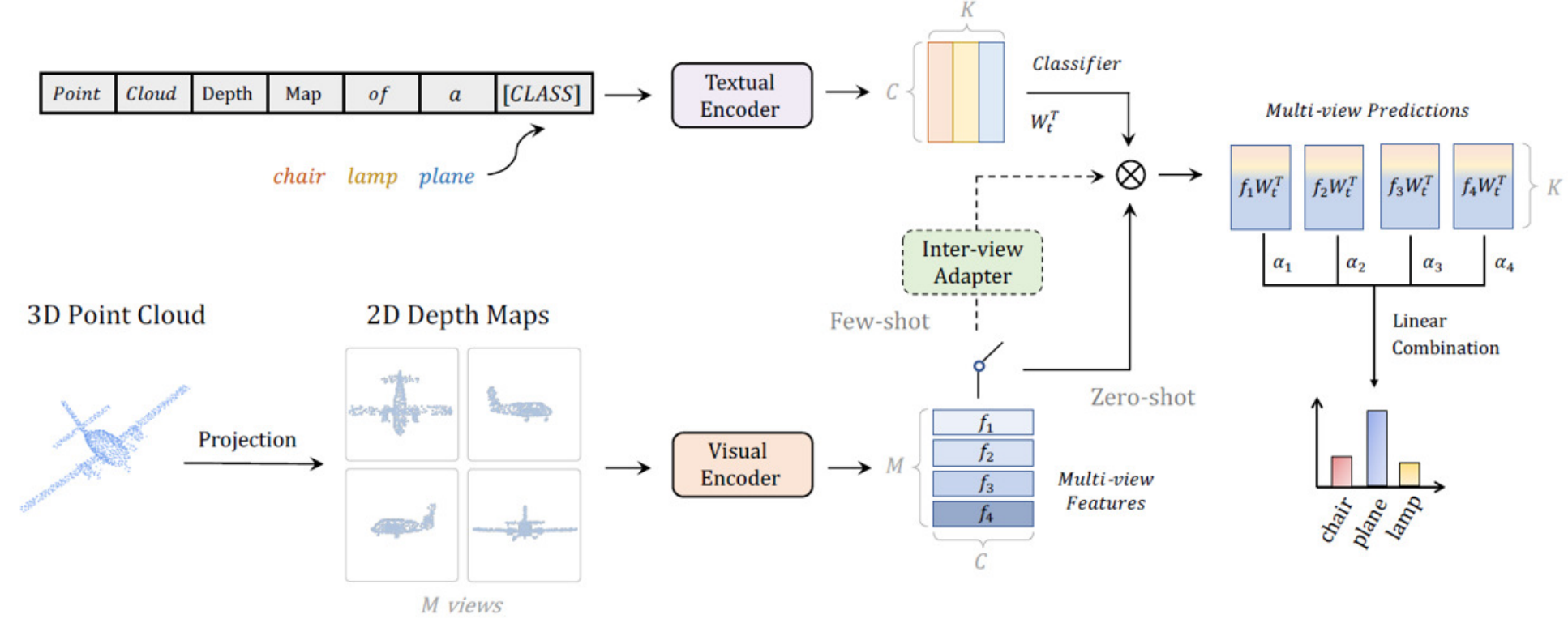}
\caption{PointCLIP\cite{zhang2022pointclip} architecture}
\label{f12}
\end{figure}

PointCLIP proposed by Zhang et al. \cite{zhang2022pointclip} learns point clouds based on pre-trained CLIP. Encoding the point cloud by projecting it into a multi-view depth map without rendering, enables zero-shot recognition by transferring the 2D pretrained knowledge to the 3D domain. And an inter-view adapter is designed to better extract global features. The network architecture is shown in Fig.~\ref{f12}.

CrossPoint \cite{afham2022crosspoint} achieves 2D-to-3D correspondence by maximizing the point cloud and the corresponding rendered 2D image in invariant space and keeping the point cloud unchanged across transformations.

Summary: Compared with the multi-view-based method and the voxel-based method, the point cloud-based method directly processes the original points and can make full use of the point cloud information. Therefore, the point cloud-based method is also a future research direction, and the transformer-based method will be more widely used in the future.

\section{Evaluation}\label{sec4}

The evaluation index is used to evaluate the performance of the point cloud classification method. The accuracy, space complexity, execution time, etc. are the evaluation indexes of the method, and the accuracy is the key index for evaluating various methods. Generally, accuracy (Acc), precision (Pre), recall (Rec), and intersection-over-union (IoU) are used to evaluate the accuracy of the method.

$\bullet$ Accuracy refers to the ratio of the number of correctly predicted samples to the total number of predicted samples.

$\bullet$ The accuracy rate refers to the proportion of the true positive class that is predicted as a positive class.

$\bullet$ Recall refers to the ratio of samples predicted to be positive classes to the total number of true positive classes.

$\bullet$ The intersection ratio refers to the ratio of the intersection and union of the predicted value and the true value. 

The accuracy, precision, recall, and intersection ratio can be calculated by the following formulas: TP is the sample that is predicted to be a positive class and is actually a positive class, TN is a sample that is predicted to be a positive class and is actually a negative class, and FP is the sample. The samples that are predicted to be negative classes are actually negative classes, and FN is the samples that are predicted to be negative classes and are actually positive classes. Suppose there are N classes:

The accuracy of the i-th class:
\begin{equation}\label{e1}
A c c_{i}=\frac{T P_{i}+T N_{i}}{T P_{i}+T N_{i}+F P_{i}+F N_{i}}
\end{equation}

The precision of the i-th class:
\begin{equation}\label{e2}
\operatorname{Pre}_{i}=\frac{T P_{i}}{T P_{i}+F P_{i}}
\end{equation}

Recall:
\begin{equation}\label{e3}
R e c=\frac{T P}{T P+F N}
\end{equation}

The intersection ratio of the i-th class:
\begin{equation}\label{e4}
I o U_{i}=\frac{T P_{i}}{T P_{i}+F P_{i}+F N_{i}}
\end{equation}

The current accuracy of point cloud classification is measured by indicators: overall accuracy (OA), average accuracy (MA), and average intersection-over-union ratio (mIoU), which are calculated as follows:
\begin{equation}\label{e5}
O A=\frac{\sum_{i=1}^{N} T P_{i}}{\sum_{i=1}^{N}\left(T P_{i}+F P_{i}\right)}
\end{equation}

\begin{equation}\label{e6}
M A=\frac{1}{N} \sum_{i=1}^{N} A c c_{i}
\end{equation}

\begin{equation}\label{e7}
m I o U=\frac{1}{N} \sum_{i=1}^{N} I o U
\end{equation}

This section summarizes the performance of the mentioned methods on different datasets in Table.~\ref{tab2}.

\begin{table*}[htbp]
\centering
\caption{Accuracy comparison of point cloud classification methods on different datasets}
\resizebox{\textwidth}{100mm}{
	
	\begin{tabular}{|ccc|cc|ccc|ccc|ccc|ccc|}
		\toprule
		\multicolumn{4}{|c}{\multirow{3}[4]{*}{Methods}} & \multirow{3}[4]{*}{Year} & \multicolumn{12}{c|}{\textcolor[rgb]{ .267,  .329,  .416}{Accuracy/\% on different datasets}} \\
		\cmidrule{6-17}    \multicolumn{4}{|c}{}         &       & \multicolumn{3}{c|}{ModelNet 40} & \multicolumn{3}{c|}{ModelNet 10} & \multicolumn{3}{c|}{ScanObjectNN} & \multicolumn{3}{c|}{ShapeNet} \\
		\multicolumn{4}{|c}{}         &       & OA    & MA    & mIoU  & OA    & MA    & mIoU  & OA    & MA    & mIoU  & OA    & MA    & mIoU \\
		\midrule
		\multicolumn{3}{|c|}{\multirow{6}[2]{*}{Multi-view based methods}} & MVCNN\cite{2015Multi}      & 2016  & 90.10 & 78.90 & -     & -     & -     & -     & -     & -     & -     & -     & -     & - \\
		\multicolumn{3}{|c|}{} & GIFT\cite{2016GIFT}  & 2016  & 83.10 & 81.94 & -     & 92.35 & 91.12 & -     & -     & -     & -     & -     & -     & - \\
		\multicolumn{3}{|c|}{} & GCVNN\cite{2018GVCNN} & 2018  & 93.10 & 79.70 & -     & -     & -     & -     & -     & -     & -     & -     & -     & - \\
		\multicolumn{3}{|c|}{} & MHBN\cite{yu2018multi}  & 2018  & 94.91 & -     & -     & 92.93 & -     & -     & -     & -     & -     & -     & -     & - \\
		\multicolumn{3}{|c|}{} & RCPCNN\cite{wang2019dominant} & 2019  & 93.80 & -     & -     & -     & -     & -     & -     & -     & -     & -     & -     & - \\
		\multicolumn{3}{|c|}{} & MVTN\cite{hamdi2021mvtn}  & 2021  & 92.00 & 93.80 & -     & -     & -     & -     & 82.80 & -     & -     & -     & -     & - \\
		\midrule
		\multicolumn{3}{|c|}{\multirow{9}[2]{*}{Voxel-based methods}} & VoxNet\cite{maturana2015voxnet} & 2015  & 85.90 & 83.00 & -     & -     & 92.00 & -     & -     & -     & -     & -     & -     & - \\
		\multicolumn{3}{|c|}{} & 3D shapeNet\cite{wu20153d} & 2015  & 84.70 & 77.30 & -     & -     & 83.50 & -     & -     & -     & -     & -     & -     & - \\
		\multicolumn{3}{|c|}{} & FPNN\cite{li2016fpnn}  & 2016  & 87.50 & -     & -     & -     & -     & -     & -     & -     & -     & -     & -     & - \\
		\multicolumn{3}{|c|}{} & OctNet\cite{riegler2017octnet} & 2017  & 86.50 & -     & -     & 90.90 & -     & -     & -     & -     & -     & -     & -     & - \\
		\multicolumn{3}{|c|}{} & O-CNN\cite{wang2017cnn} & 2017  & 90.60 & -     & 85.90 & -     & -     & -     & -     & -     & -     & -     & -     & - \\
		\multicolumn{3}{|c|}{} & Kd-Net\cite{klokov2017escape} & 2017  & 91.80 & 88.50 & -     & 94.00 & 93.50 & 77.20 & -     & -     & -     & -     & -     & 82.30 \\
		\multicolumn{3}{|c|}{} & 3D Context Net\cite{zeng20183dcontextnet} & 2018  & -     & -     & -     & -     & -     & -     & -     & -     & -     & -     & -     & - \\
		\multicolumn{3}{|c|}{} & MSNet\cite{wang2018msnet} & 2018  & -     & -     & -     & -     & -     & -     & -     & -     & -     & -     & -     & - \\
		\multicolumn{3}{|c|}{} & VV-Net\cite{meng2019vv} & 2019  & -     & -     & -     & -     & -     & -     & -     & -     & -     & -     & -     & 87.40 \\
		\midrule
		\multicolumn{1}{|c|}{\multirow{46}[14]{*}{Point cloud-based method}} & \multicolumn{1}{c|}{\multirow{33}[8]{*}{Local feature aggregation}} & \multirow{8}[2]{*}{Point-by-point methods} & PointNet++\cite{qi2017pointnet++} & 2017  & 91.90 & -     & -     & -     & -     & -     & 77.90 & 75.40 & -     & -     & -     & 85.10 \\
		\multicolumn{1}{|c|}{} & \multicolumn{1}{c|}{} &       & PointNeXt\cite{qian2022pointnext} & 2022  & - & - & -     & -     & -     & -     & 87.70     & 85.80     & -     & -     & -     & 87.20 \\
		\multicolumn{1}{|c|}{} & \multicolumn{1}{c|}{} &       & PointWeb\cite{zhao2019pointweb} & 2019  & 92.30 & 89.40 & -     & -     & -     & -     & -     & -     & -     & -     & -     & - \\
		\multicolumn{1}{|c|}{} & \multicolumn{1}{c|}{} &       & RandLA-Net\cite{hu2020randla} & 2020  & -     & -     & -     & -     & -     & -     & -     & -     & -     & -     & -     & - \\
		\multicolumn{1}{|c|}{} & \multicolumn{1}{c|}{} &       & SO-Net\cite{li2018so} & 2018  & 90.80 & 87.30 & -     & 94.10 & 93.90 & -     & -     & -     & -     & -     & -     & 84.60 \\
		\multicolumn{1}{|c|}{} & \multicolumn{1}{c|}{} &       & GDANet\cite{xu2021learning} & 2021  & 93.80 & -     & -     & -     & -     & -     & 88.50 & -     & -     & -     & -     & 86.50 \\
		\multicolumn{1}{|c|}{} & \multicolumn{1}{c|}{} &       & SimpleView\cite{goyal2020revisiting} & 2020  & 93.00 & -     & -     & -     & -     & -     & 80.50 & -     & -     & -     & -     & - \\
		\multicolumn{1}{|c|}{} & \multicolumn{1}{c|}{} &       & PintSCNet\cite{chen2021pointscnet} & 2021  & 93.70 & -     & -     & -     & -     & -     & -     & -     & -     & -     & -     & - \\
		\multicolumn{1}{|c|}{} & \multicolumn{1}{c|}{} &       & PointMLP\cite{ma2022rethinking} & 2022  & 94.50 & 91.40 & -     & -     & -     & -     & 85.40 & 83.90 & -     & -     & -     & - \\
		\multicolumn{1}{|c|}{} & \multicolumn{1}{c|}{} &       & RepSurf\cite{ran2022surface} & 2022  & 94.70 & 91.70 & -     & -     & -     & -     & 84.60 & 81.90 & -     & -     & -     & - \\
		\cmidrule{3-3}    \multicolumn{1}{|c|}{} & \multicolumn{1}{c|}{} & \multirow{12}[2]{*}{Convolution based methods} & PCNN\cite{atzmon2018point}  & 2018  & 92.30 & -     & -     & 94.90 & -     & -     & -     & -     & -     & -     & -     & - \\
		\multicolumn{1}{|c|}{} & \multicolumn{1}{c|}{} &       & RS-CNN\cite{liu2019relation} & 2019  & 93.60 & -     & -     & -     & -     & -     & -     & -     & -     & -     & -     & 86.20 \\
		\multicolumn{1}{|c|}{} & \multicolumn{1}{c|}{} &       & DNNSP\cite{wang2018deep} & 2018  & -     & -     & -     & -     & -     & -     & -     & -     & -     & -     & -     & - \\
		\multicolumn{1}{|c|}{} & \multicolumn{1}{c|}{} &       & A-CNN\cite{komarichev2019cnn} & 2019  & 92.60 & 90.30 & -     & 95.50 & 95.30 & -     & -     & -     & -     & -     & -     & 86.10 \\
		\multicolumn{1}{|c|}{} & \multicolumn{1}{c|}{} &       & RP-Net\cite{ran2021learning} & 2021  & 94.10 & -     & -     & -     & -     & -     & -     & -     & -     & -     & -     & - \\
		\multicolumn{1}{|c|}{} & \multicolumn{1}{c|}{} &       & SCN\cite{xie2018attentional}   & 2018  & 90.00 & -     & -     & -     & -     & -     & -     & -     & -     & -     & -     & 84.60 \\
		\multicolumn{1}{|c|}{} & \multicolumn{1}{c|}{} &       & PointCNN\cite{li2018pointcnn} & 2018  & 92.20 & 88.10 & -     & -     & -     & -     & 87.90 & -     & -     & -     & -     & - \\
		\multicolumn{1}{|c|}{} & \multicolumn{1}{c|}{} &       & PointCove\cite{wu2019pointconv} & 2019  & 92.50 & -     & -     & -     & -     & -     & -     & -     & -     & -     & -     & - \\
		\multicolumn{1}{|c|}{} & \multicolumn{1}{c|}{} &       & MCCNN\cite{hermosilla2018monte} & 2018  & 90.90 & -     & -     & -     & -     & -     & -     & -     & -     & -     & -     & - \\
		\multicolumn{1}{|c|}{} & \multicolumn{1}{c|}{} &       & SpiderCNN\cite{xu2018spidercnn} & 2018  & 92.40 & -     & -     & -     & -     & -     & 73.70 & 69.80 & -     & -     & -     & 85.30 \\
		\multicolumn{1}{|c|}{} & \multicolumn{1}{c|}{} &       & InterpCNN\cite{mao2019interpolated} & 2019  & 93.00 & -     & -     & -     & -     & -     & -     & -     & -     & -     & -     & 86.30 \\
		\multicolumn{1}{|c|}{} & \multicolumn{1}{c|}{} &       & SPHNet\cite{poulenard2019effective} & 2019  & 87.70 & -     & -     & -     & -     & -     & -     & -     & -     & -     & -     & - \\
		\multicolumn{1}{|c|}{} & \multicolumn{1}{c|}{} &       & SC-CNN\cite{wang2022learning} & 2022  & 93.8 & -     & -     & 96.4     & -     & -     & -     & -     & -     & -     & -     & 86.40 \\
		\cmidrule{3-3}    \multicolumn{1}{|c|}{} & \multicolumn{1}{c|}{} & \multirow{8}[2]{*}{Graph-based methods} & ECC\cite{simonovsky2017dynamic}   & 2017  & -     & 83.20 & -     & -     & 90.00 & -     & -     & -     & -     & -     & -     & - \\
		\multicolumn{1}{|c|}{} & \multicolumn{1}{c|}{} &       & SpecGCN\cite{wang2018local} & 2018  & 91.50 & -     & -     & -     & -     & -     & -     & -     & -     & -     & -     & 84.60 \\
		\multicolumn{1}{|c|}{} & \multicolumn{1}{c|}{} &       & Grid-GCN\cite{xu2020grid} & 2020  & 93.10 & 91.30 & -     & 97.50 & 97.40 & -     & -     & -     & -     & -     & -     & - \\
		\multicolumn{1}{|c|}{} & \multicolumn{1}{c|}{} &       & PointView-GCN\cite{mohammadi2021pointview} & 2021  & 95.40 & -     & -     & -     & -     & -     & 85.50 & -     & -     & -     & -     & - \\
		\multicolumn{1}{|c|}{} & \multicolumn{1}{c|}{} &       & DGCNN\cite{wang2019dynamic} & 2019  & 92.20 & 90.20 & -     & -     & -     & -     & 86.20 & -     & -     & -     & -     & - \\
		\multicolumn{1}{|c|}{} & \multicolumn{1}{c|}{} &       & LDGCNN\cite{zhang2019linked} & 2019  & 92.90 & 90.30 & -     & -     & -     & -     & -     & -     & -     & -     & -     & - \\
		\multicolumn{1}{|c|}{} & \multicolumn{1}{c|}{} &       & PointNGCNN\cite{lu2020pointngcnn} & 2020  & 92.80 & -     & -     & -     & -     & -     & -     & -     & -     & -     & -     & 85.60 \\
		
		\cmidrule{3-3}    \multicolumn{1}{|c|}{} & \multicolumn{1}{c|}{} & \multirow{5}[2]{*}{attention-based methods} & PAT\cite{yang2019modeling}   & 2019  & 91.70 & -     & -     & -     & -     & -     & -     & -     & -     & -     & -     & - \\
		\multicolumn{1}{|c|}{} & \multicolumn{1}{c|}{} &       & FPA/GAU\cite{li2018pyramid} & 2018  & -     & -     & -     & -     & -     & -     & -     & -     & -     & -     & -     & - \\
		\multicolumn{1}{|c|}{} & \multicolumn{1}{c|}{} &       & LSANet\cite{chen2019lsanet} & 2019  & 92.30 & 89.20 & -     & -     & -     & -     & -     & -     & -     & -     & -     & 83.20 \\
		\multicolumn{1}{|c|}{} & \multicolumn{1}{c|}{} &       & GACNet\cite{wang2019graph} & 2019  & -     & -     & -     & -     & -     & -     & -     & -     & -     & -     & -     & - \\
		\multicolumn{1}{|c|}{} & \multicolumn{1}{c|}{} &       & GAPointNet\cite{chen2021gapointnet} & 2021  & 92.40 & 89.70 & -     & -     & -     & -     & -     & -     & -     & -     & -     & - \\
		\cmidrule{2-3}    \multicolumn{1}{|c|}{} & \multicolumn{1}{c|}{\multirow{12}[5]{*}{Global feature aggregation}} & \multirow{6}[2]{*}{Transformer-based methods} & Point Transformer\cite{engel2021point} & 2021  & 92.80 & -     & -     & -     & -     & -     & -     & -     & -     & 85.90 & -     & - \\
		\multicolumn{1}{|c|}{} & \multicolumn{1}{c|}{} &       & Visual Transformer\cite{wu2020visual} & 2020  & -     & -     & -     & -     & -     & -     & -     & -     & -     & -     & -     & - \\
		\multicolumn{1}{|c|}{} & \multicolumn{1}{c|}{} &       & PCT\cite{guo2021pct}   & 2021  & 93.20 & -     & -     & -     & -     & -     & -     & -     & -     & -     & -     & - \\
		\multicolumn{1}{|c|}{} & \multicolumn{1}{c|}{} &       & 3DMedPT\cite{yu20213d} & 2021  & 93.40 & -     & -     & -     & -     & -     & -     & -     & -     & -     & -     & - \\
		\multicolumn{1}{|c|}{} & \multicolumn{1}{c|}{} &       & Point-BERT\cite{yu2022point} & 2021  & 93.80 & -     & -     & -     & -     & -     & 83.07 & -     & -     & -     & -     & - \\
		\multicolumn{1}{|c|}{} & \multicolumn{1}{c|}{} &       & Point-TnT\cite{berg2022points} & 2022  & 92.60 & -     & -     & -     & -     & -     & 84.60 & 82.60 & -     & -     & -     & - \\
		\multicolumn{1}{|c|}{} & \multicolumn{1}{c|}{} &       & Point-MAE\cite{pang2022masked} & 2022  & 93.80 & -     & -     & -     & -     & -     & 85.18 & - & -     & -     & -     & 86.10 \\
		\cmidrule{3-3}    \multicolumn{1}{|c|}{} & \multicolumn{1}{c|}{} & \multirow{3}[2]{*}{Global module-based methods} & PointNet\cite{qi2017pointnet} & 2017  & 89.20 & 86.20 & -     & -     & -     & -     & 68.20 & 63.40 & -     & -     & -     & 83.70 \\
		\multicolumn{1}{|c|}{} & \multicolumn{1}{c|}{} &       & PointASNL\cite{yan2020pointasnl} & 2020  & 93.20 & -     & -     & 95.90 & -     & -     & -     & -     & -     & -     & -     & - \\
		\multicolumn{1}{|c|}{} & \multicolumn{1}{c|}{} &       & CurveNet\cite{xiang2021walk} & 2021  & 94.20 & -     & -     & 96.30 & -     & -     & -     & -     & -     & -     & -     & 86.80 \\
		\multicolumn{1}{|c|}{} & \multicolumn{1}{c|}{} &       & DeepGCNs\cite{li2021deepgcns} & 2021  & 93.20 & 90.30 & -     & -     & -     & -     & -     & -     & -     & -     & -     & - \\
		\cmidrule{3-3}    \multicolumn{1}{|c|}{} & \multicolumn{1}{c|}{} & \multirow{4}[2]{*}{RNN or LSTM based methods} & 3DCNN-DQN-RNN\cite{liu20173dcnn} & 2017  & -     & -     & -     & -     & -     & -     & -     & -     & -     & -     & -     & - \\
		\multicolumn{1}{|c|}{} & \multicolumn{1}{c|}{} &       & RSNet\cite{huang2018recurrent} & 2018  & -     & -     & -     & -     & -     & -     & -     & -     & -     & -     & -     & - \\
		\multicolumn{1}{|c|}{} & \multicolumn{1}{c|}{} &       & 3P-RNN\cite{ye20183d} & 2018  & -     & -     & -     & -     & -     & -     & -     & -     & -     & -     & -     & - \\
		\multicolumn{1}{|c|}{} & \multicolumn{1}{r|}{} &       & Point2Sequence\cite{liu2019point2sequence} & 2019  & 92.60 & 90.40 & -     & 95.30 & 95.10 & -     & -     & -     & -     & -     & -     & - \\
		\midrule
		\multicolumn{3}{|c|}{\multirow{4}[2]{*}{Polymorphic Fusion based methods}} & PointGrid\cite{le2018pointgrid} & 2018  & 92.00 & 88.90 & -     & -     & -     & -     & -     & -     & -     & 86.10 & 80.50 & - \\
		\multicolumn{3}{|c|}{} & PVT\cite{zhang2021pvt}   & 2021  & 94.00 & -     & -     & -     & -     & -     & -     & -     & -     & -     & -     & 86.60 \\
		\multicolumn{3}{|c|}{} & PointCLIP\cite{zhang2022pointclip} & 2022  & 94.05 & -     & -     & -     & -     & -     & -     & -     & -     & -     & -     & - \\
		\multicolumn{3}{|c|}{} & CrossPoint\cite{afham2022crosspoint} & 2022  & 94.90 & -     & -     & -     & -     & -     & 79.00 & -     & -     & -     & -     & 85.50 \\
		\bottomrule
\end{tabular}}%
\label{tab2}%
\end{table*}%

\section{Future trends}\label{sec5}

With the development of today's technology, the demand for point cloud classification methods in various fields is getting higher and higher. There are many methods of point cloud classification, researchers are constantly proposing new methods to improve accuracy and efficiency. This continues to drive the development of 3D applications. However, the existing methods still have different limitations. This section will summarize the problems in the current deep learning-based point cloud classification methods and prospect the future research directions, details as follows:

$\bullet$ Some of the current point cloud classification methods focus on improving accuracy, while others focus on improving efficiency. We need to solve the problem of "how to achieve high accuracy while being efficient?" 

$\bullet$ In the practical application of 3D, the information structure of outdoor scenes is complex and changeable. Although some methods have been applied in outdoor scenes, the efficiency and accuracy need to be further improved. Therefore, future research direction should focus more on the network for outdoor scenes.

$\bullet$ From the above data, the method based on the original point cloud has certain advantages in algorithm performance. But the network model is more complicated, because the input original point cloud data has information integrity, therefore simple point cloud-based methods are a future research trend. 

$\bullet$ Most of the current methods are aimed at improving rather than changing, so the innovative type of methods needs our continued research and discussion. 

$\bullet$ In the network model we will propose in the future, we should pay attention to the optimization of the network architecture, which can reduce the computational complexity and memory usage while facing the complex and irregular point cloud.

\section{Conclusion}\label{sec6}

This paper provides a comprehensive survey and discussion of deep learning-based point cloud classification methods in recent years. 

First, we introduced the point cloud and its application in the introduction and discussed the characteristics and processing difficulties of the point cloud. In the second section, the 3D data were introduced, and the commonly used 3D data representations, point cloud data storage formats and point cloud classification datasets are summarized. Building on the previous sections, we comprehensively review deep learning-based point cloud classification methods, classifying these methods into four broad categories: multi-view-based methods, voxel-based methods,  point cloud-based methods, and polymorphic fusion-based methods. And then we compare the performance of existing methods. Finally, this paper points out the problems of the current method and prospects the future research direction.

\section*{Acknowledgment}

This work is supported by Xinjiang University School-Enterprise Joint Project.

\bibliography{reference}
\end{document}